\documentclass[sigconf]{acmart}
\usepackage{graphicx}
\usepackage{multirow}
\usepackage{array}
\usepackage[table]{xcolor}
\AtBeginDocument{%
  }

\setcopyright{acmlicensed}
\copyrightyear{2018}
\acmYear{2018}
\acmDOI{XXXXXXX.XXXXXXX}
\acmConference[Conference acronym 'XX]{Make sure to enter the correct
  conference title from your rights confirmation email}{June 03--05,
  2018}{Woodstock, NY}
\acmISBN{978-1-4503-XXXX-X/2018/06}

\begin{document}

\title{LEGS: Laplacian-Enhanced Gaussian Splatting with a Nonlinear Weighted Loss}


\author{Yongfei Guo, Qizhou Huo, Xuan Sun, Yuanhao Gong}
\affiliation{%
  \institution{Changchun Institute of Optics, Fine Mechanics and Physics, Chinese Academy of Sciences, Changchun, China \and
  	University of Chinese Academy of Sciences, Beijing, China}
  \city{}
  \country{}}

%
%
\renewcommand{\shortauthors}{Yongfei et al.}

\begin{abstract}

3D Gaussian Splatting (3DGS) has become an efficient explicit representation for radiance field reconstruction and real-time novel view synthesis. However, its standard photometric loss treats flat and structure-rich regions similarly, which may limit the recovery of sharp contours and fine details. Edge-Guided Gaussian Splatting (EGGS) improves structure awareness through edge-guided weighting, but mainly relies on first-order gradient responses and linear weighting. In this paper, we propose LEGS, a Laplacian-Enhanced Gaussian Splatting method with a nonlinearly weighted loss. LEGS replaces first-order gradient guidance with second-order Laplacian structural guidance and maps the normalized Laplacian response into pixel-wise weights through nonlinear response-to-weight functions. The proposed loss improves structure-aware Gaussian optimization while keeping the original 3DGS rendering pipeline unchanged. Experiments on the full Tanks\&Temples and Mip-NeRF360 datasets show that LEGS improves peak signal-to-noise ratio (PSNR) by up to 1.68 dB over 3DGS and up to 0.52 dB over EGGS. Incorporating the proposed second-order nonlinear weighting strategy into FastGS and FasterGS further improves PSNR by up to 1.69 dB, demonstrating its effectiveness as a general loss-level extension for Gaussian Splatting pipelines with potential applications in AR/VR, immersive visualization, and real-time 3D content generation.
\end{abstract}

\ccsdesc[500]{Computing methodologies~Computational photography; Reconstruction; Rasterization}

\keywords{Gaussian, Splatting, Radiance, Laplacian, Nonlinear Weighted Loss}


\begin{teaserfigure}
	\centering
	\makebox[\textwidth][c]{%
		\includegraphics[width=0.75\textwidth]{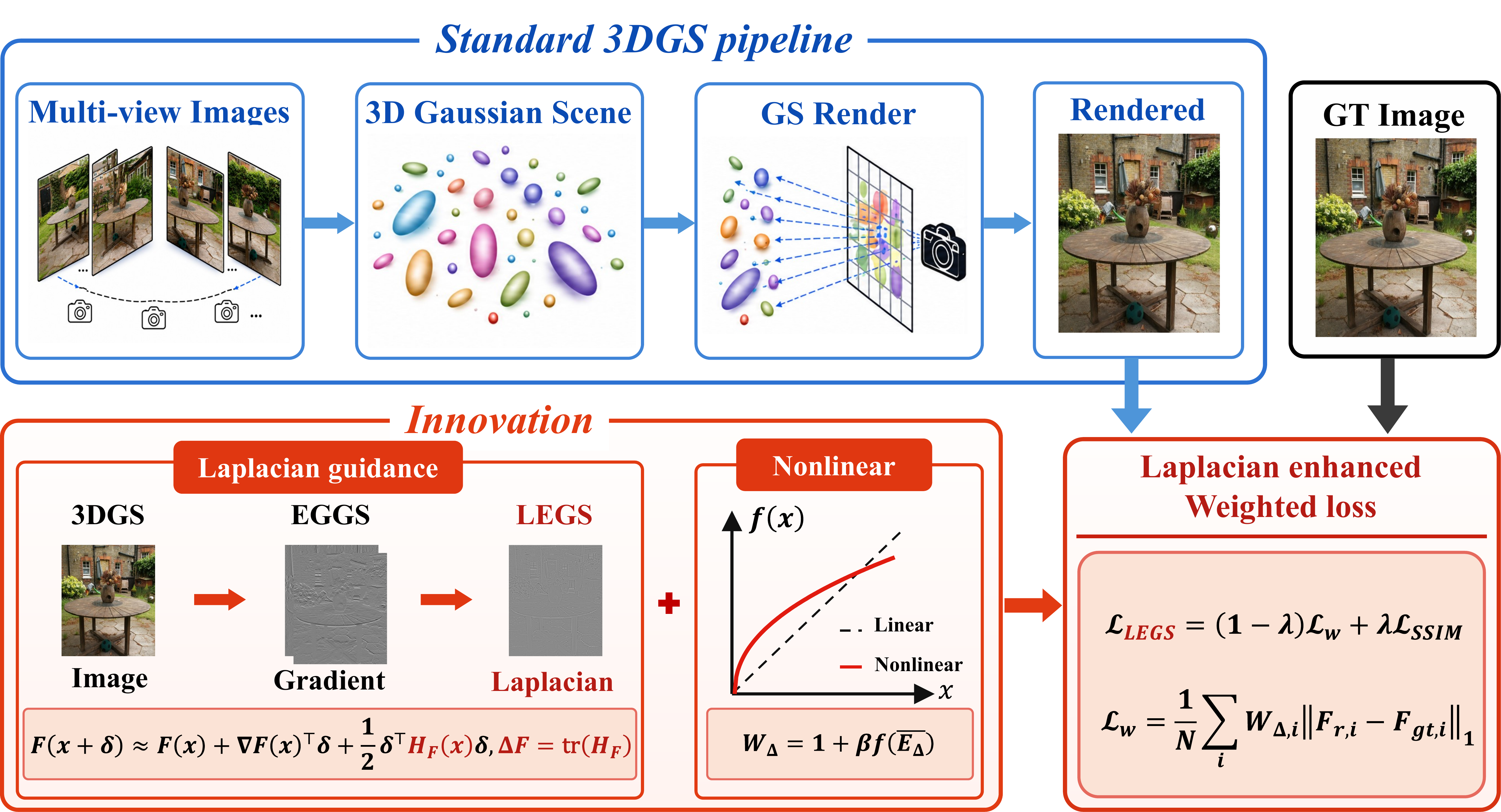}
	}
	\caption{Overview of LEGS. The standard 3DGS pipeline is kept unchanged, while Laplacian structural guidance and nonlinear weights are introduced only in the reconstruction loss.}
	\Description{An overview diagram of the LEGS framework. The top part shows the unchanged standard 3DGS pipeline from multi-view images to a 3D Gaussian scene, differentiable Gaussian splatting, and a rendered view. The bottom part highlights two proposed components, Laplacian structural guidance and nonlinear response-to-weight mapping, which are used to form a Laplacian-enhanced weighted reconstruction loss.}
	\label{fig:teaser}
\end{teaserfigure}


\maketitle

\section{Introduction}

Radiance field reconstruction from multi-view images is a fundamental problem in computer vision and graphics, with broad applications in novel view synthesis, immersive visualization, AR/VR, and web-based 3D content generation~\cite{NeRF,Mip-NeRF360,Instant-NGP,Plenoxels}. Recently, 3D Gaussian Splatting (3DGS) has become an effective explicit representation for this task, enabling high-quality radiance field reconstruction and real-time rendering~\cite{3DGS,Survey1,Survey2}. Existing 3DGS variants improve rendering quality, efficiency, and compactness through anti-aliasing splatting, structured Gaussian representations, or Gaussian compression~\cite{Mip-Splatting,scaffoldgs,LightGaussian}. In contrast, this work focuses on loss-level structural guidance without changing the Gaussian representation or rendering pipeline.

Despite its success, 3DGS optimization is still mainly driven by photometric reconstruction losses, which treat flat, weakly textured, and structure-rich regions similarly. As a result, important contours, corners, thin structures, and high-frequency details may not receive sufficient emphasis. Edge-Guided Gaussian Splatting (EGGS)~\cite{EGGS} addresses this issue by assigning larger reconstruction weights to edge pixels, showing that loss-level edge guidance can improve Gaussian Splatting without modifying the rendering pipeline. However, EGGS relies on first-order gradient magnitude and linear response-to-weight mapping, which may not fully capture second-order
variations~\cite{EdgeGaussians}.

In this paper, we propose \textbf{LEGS}, a Laplacian-Enhanced Gaussian Splatting method with a nonlinear weighted loss. LEGS replaces first-order gradient guidance with the absolute Laplacian response of input images, which provides a compact second-order cue for structure-sensitive regions. Nonlinear response-to-weight mappings are further introduced to control the contribution of different structural response levels. LEGS improves structure-aware Gaussian optimization while keeping the original 3DGS rendering and optimization pipeline unchanged.

\subsection{3D Gaussian Splatting}

3D Gaussian Splatting represents a radiance field using a set of anisotropic Gaussian primitives. Each primitive is associated with a 3D position, covariance, opacity, and view-dependent color parameters. Given a camera view, the 3D Gaussians are projected onto the image plane and composited through differentiable alpha blending. This explicit representation enables efficient optimization and real-time rendering, making 3DGS attractive for high-quality novel view synthesis.

Let $U$ and $F$ denote the rendered image and the supervision image, respectively. The standard 3DGS reconstruction loss is commonly written as
\begin{equation}
	L_{\mathrm{3DGS}}
	=
	(1-\lambda)\|U-F\|_1
	+
	\lambda D_{\mathrm{SSIM}}(U,F),
\end{equation}
where $D_{\mathrm{SSIM}}$ is the SSIM-based distance and $\lambda$ balances the two terms. Although effective, this objective does not explicitly distinguish structure-rich pixels from flat ones, which may weaken the reconstruction of sharp boundaries and fine details.

\subsection{Edge-Guided Gaussian Splatting}

Edges and local structures are important visual cues, and edge-preserving filtering studies show that preserving structural discontinuities helps avoid over-smoothing and maintain perceptual fidelity~\cite{SWF, SWGF,QLF,SWBF}. In radiance field reconstruction, such cues often correspond to object boundaries, depth discontinuities, and fine details. EGGS introduces them into Gaussian Splatting through a gradient-based weight:
\begin{equation}
	W_{g}(u,v)=1+\beta\|\nabla F(u,v)\|_p,
\end{equation}
where $(u,v)$ is the pixel coordinate, $\nabla$ is the image gradient operator, and $\beta$ controls edge amplification.

This loss-level design is simple and effective, but its guidance is based on first-order image variation. Gradient magnitude mainly captures local intensity slopes and may be insufficient for second-order structures such as corners, thin contours, and high-frequency details. Moreover, a linear response-to-weight mapping may not adapt well to different scenes and response distributions.

\subsection{From First-Order to Second-Order}

To introduce second-order structural guidance, we use the Laplacian response:
\begin{equation}
	E_{l}(u,v)=\|\Delta F(u,v)\|_p,
\end{equation}
where $\Delta$ denotes the image Laplacian operator and the norm converts the signed response into a non-negative magnitude. Compared with gradient magnitude, the Laplacian provides a compact structure-sensitive cue for contours, corners, thin structures, and high-frequency local details.

We further map the Laplacian response into loss weights. Since Laplacian responses can be unevenly distributed, a nonlinear response-to-weight mapping offers more flexible control over weak, medium, and strong structural responses during Gaussian optimization.

\subsection{Motivations and Contributions}

The main motivation of this work is to improve the structure awareness of Gaussian Splatting with a lightweight loss-level design. LEGS replaces
first-order gradient guidance with second-order Laplacian structural guidance and further reshapes the response through nonlinear weighting,
while keeping the original 3DGS representation and rendering pipeline unchanged.

The main contributions are summarized as follows:
\begin{itemize}
	\item We propose \textbf{LEGS}, a Laplacian-enhanced Gaussian Splatting method that introduces second-order structural guidance into radiance
	field reconstruction.
	
	\item We design a nonlinearly Laplacian-weighted loss to flexibly control the contribution of structure-rich pixels during Gaussian optimization.
	
	\item We evaluate LEGS on T\&T and Mip-NeRF360, analyze the weighting parameter $\beta$ and nonlinear mappings, and further verify the generality of the proposed loss on FastGS and FasterGS under different Gaussian Splatting pipelines.
\end{itemize}

\section{Laplacian-Enhanced Gaussian Splatting}

Fig.~\ref{fig:teaser} illustrates the overview of the proposed LEGS framework. LEGS keeps the standard 3DGS pipeline unchanged, including the Gaussian representation, differentiable Gaussian splatting renderer, and densification strategy, while introducing a Laplacian-enhanced nonlinear weighted loss for structure-aware optimization.

\subsection{Overview}

Given a set of multi-view training images $\mathcal{F}=\{F^i\}_{i=1}^{N}$, LEGS first computes a Laplacian structural response map for each input view. Since the response map depends only on the observed image rather than the optimized radiance field, it can be pre-computed and reused as a fixed structural prior throughout optimization.

For each view, the Laplacian response is normalized to a common range and then passed through a nonlinear response-to-weight mapping. The mapped response is used to generate a pixel-wise weight map, which reweights the photometric reconstruction term. In this way, pixels with stronger second-order structural variations contribute more to Gaussian optimization, while the original 3DGS rendering process remains unchanged.

\subsection{Laplacian Structural Guidance}

For the $i$-th input image $F^i$, we compute the Laplacian structural response as
\begin{equation}
	E_{\Delta}^{i}(u,v)
	=
	\left\|
	\Delta F^i(u,v)
	\right\|_p,
\end{equation}
where $\Delta$ denotes the image Laplacian operator, and $\|\cdot\|_p$ computes the magnitude of the Laplacian response. For RGB images, $\Delta F^i(u,v)$ is regarded as a channel-wise response vector.

To make the weighting factor comparable across different views and scenes, the response map is normalized to $[0,1]$:
\begin{equation}
	\hat{E}_{\Delta}^{i}(u,v)
	=
	\frac{
		E_{\Delta}^{i}(u,v)-E_{\Delta,\min}^{i}
	}{
		E_{\Delta,\max}^{i}-E_{\Delta,\min}^{i}+\epsilon
	},
\end{equation}
where $E_{\Delta,\min}^{i}$ and $E_{\Delta,\max}^{i}$ are the minimum and maximum values of $E_{\Delta}^{i}$ over the image domain, and $\epsilon$ is a small constant for numerical stability. The normalized response $\hat{E}_{\Delta}^{i}$ provides a compact second-order structural cue for the following weighted loss design.

\subsection{Nonlinear Laplacian-Weighted Loss}

\begin{figure}[htbp]
	\centering
	\includegraphics[width=0.5\columnwidth]{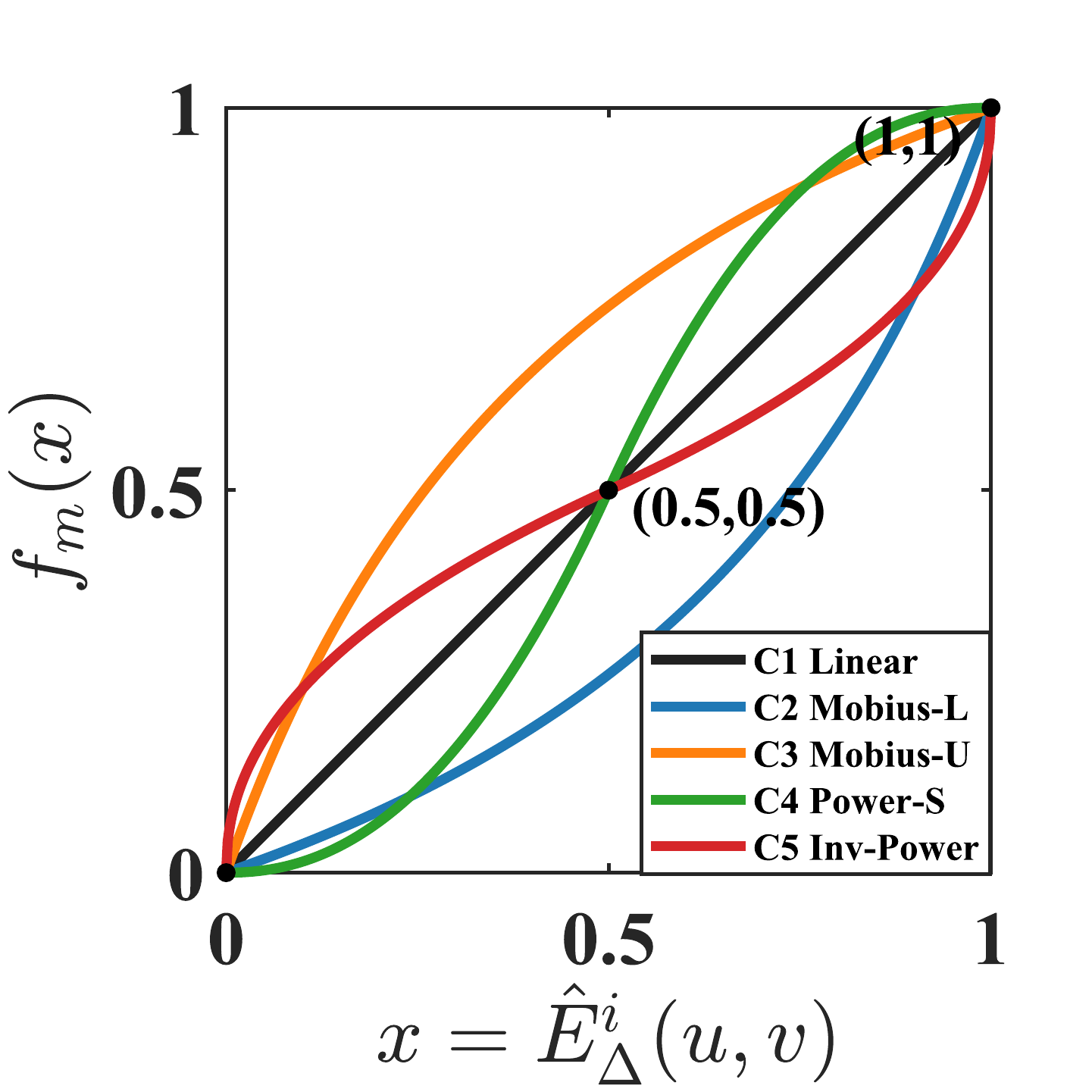}
	\caption{Nonlinear response-to-weight mapping functions. C1 is the linear reference, while C2--C5 reshape the normalized Laplacian response with different nonlinear behaviors.}
	\Description{A plot showing five response-to-weight mapping curves. C1 is linear, while C2 to C5 are nonlinear mappings with different shapes.}
	\label{fig:nonlinear_mapping}
\end{figure}

A linear response-to-weight mapping is simple, but different Laplacian response levels may not contribute equally to Gaussian optimization. Therefore, we introduce a nonlinear mapping function to reshape the normalized Laplacian response:
\begin{equation}
	\tilde{E}_{\Delta,m}^{i}(u,v)
	=
	f_m\left(\hat{E}_{\Delta}^{i}(u,v)\right),
	\quad
	m=1,\ldots,5,
\end{equation}
where $f_m:[0,1]\rightarrow[0,1]$ denotes the $m$-th mapping function. We evaluate five mappings, denoted as C1--C5. As shown in
Fig.~\ref{fig:nonlinear_mapping}, C1 is the linear reference, while C2--C5 reshape the normalized Laplacian response with different nonlinear behaviors. These mappings allow the loss to adjust the relative contributions of weak, medium, and strong structural responses.

The final Laplacian-enhanced weight map is defined as
\begin{equation}
	W_{\Delta,m}^{i}(u,v)
	=
	1+\beta \tilde{E}_{\Delta,m}^{i}(u,v),
\end{equation}
where $\beta>0$ controls the strength of Laplacian structural enhancement. When $\beta=0$, the weight map becomes one and the loss degenerates to the standard unweighted photometric loss.

Let $U^i$ denote the rendered image of the $i$-th view. The Laplacian-weighted photometric term is defined as
\begin{equation}
	L_{\mathrm{w}}
	=
	\frac{1}{N}
	\sum_{i=1}^{N}
	\frac{1}{|\Omega|}
	\sum_{(u,v)\in\Omega}
	W_{\Delta,m}^{i}(u,v)
	\left\|
	U^i(u,v)-F^i(u,v)
	\right\|_1,
\end{equation}
where $\Omega$ is the image domain. Following the standard reconstruction objective in 3DGS, we combine the weighted photometric term with an SSIM-based distance:
\begin{equation}
	L_{\mathrm{LEGS}}
	=
	(1-\lambda)L_{\mathrm{w}}
	+
	\lambda L_{\mathrm{ssim}},
\end{equation}
where $L_{\mathrm{ssim}}=\frac{1}{N}\sum_{i=1}^{N}D_{\mathrm{SSIM}}(U^i,F^i)$. The proposed loss only changes the supervision signal and does not modify the Gaussian renderer or the optimization pipeline. Therefore, LEGS can be implemented as a lightweight loss-level extension of 3DGS.

\section{Experiments}

We evaluate the proposed Laplacian-enhanced weighted loss through overall comparison, component analysis, and visual inspection. Specifically, we compare LEGS with representative Gaussian Splatting baselines, analyze the effects of $\beta$ and nonlinear response-to-weight mappings, and examine local structural details through visual comparisons.
\subsection{Experimental Setup}

We evaluate the proposed method on the full Train and Truck scenes from Tanks-and-Temples (T\&T)~\cite{tank} and the full Mip-NeRF360
dataset~\cite{mip360}. All reported results, including 3DGS and other baselines, are computed on the training views for controlled within-paper
comparison.

We compare LEGS with 3DGS, EGGS, and LEGS-L, a linear Laplacian-weighted variant. To examine generality, we further incorporate the proposed loss
into FastGS~\cite{FastGS} and FasterGS~\cite{FasterGS}. Unless otherwise specified, the Gaussian representation and renderer are unchanged, and
only the reconstruction loss is modified.
\subsection{Quantitative Results}

Table~\ref{tab:main_comparison} reports the main quantitative comparison. The upper block compares 3DGS, EGGS, LEGS-L, and LEGS. LEGS-L improves over 3DGS on both datasets, confirming the effectiveness of second-order Laplacian guidance. LEGS further outperforms LEGS-L and EGGS, showing that nonlinear Laplacian-weighted supervision provides stronger structure-aware optimization than linear Laplacian weighting and first-order gradient guidance.

To further examine the generality of the proposed loss, we apply LEGS to two efficient Gaussian Splatting variants, FastGS and FasterGS. As shown in the lower block of Table~\ref{tab:main_comparison}, replacing their original reconstruction losses with the proposed LEGS loss consistently improves the reconstruction quality on both T\&T and Mip-NeRF360. This indicates that LEGS is not limited to a specific Gaussian Splatting implementation, but can be used as a general loss-level extension.
\begin{table}[htbp]
	\centering
	\caption{Main quantitative comparison. Higher PSNR/SSIM and lower LPIPS are better.}
	\label{tab:main_comparison}
	\scriptsize
	\setlength{\tabcolsep}{2.0pt}
	\renewcommand{\arraystretch}{1.08}
	\resizebox{\columnwidth}{!}{%
		\begin{tabular}{c|ccc|ccc}
			\hline
			\hline
			\multirow{2}{*}{Method}
			& \multicolumn{3}{c|}{Tanks\&Temples}
			& \multicolumn{3}{c}{Mip-NeRF360} \\
			\cline{2-7}
			& PSNR$\uparrow$ & SSIM$\uparrow$ & LPIPS$\downarrow$
			& PSNR$\uparrow$ & SSIM$\uparrow$ & LPIPS$\downarrow$ \\
			\hline
			3DGS~\cite{3DGS}
			& 26.6731 & 0.8938 & 0.1508
			& 29.3242 & 0.8729 & 0.1896 \\
			
			EGGS~\cite{EGGS}
			& 27.8980 & 0.9052 & 0.1340
			& 30.4814 & 0.8911 & 0.1687 \\
			
			LEGS-L
			& 27.8820 & 0.9085 & 0.1286
			& 30.3920 & 0.8927 & 0.1667 \\
			
			\rowcolor{gray!15}
			LEGS (Ours)
			& \textbf{28.1961} & \textbf{0.9092} & \textbf{0.1269}
			& \textbf{31.0006} & \textbf{0.9010} & \textbf{0.1533} \\
			\hline
			
			FastGS
			& 25.7165 & 0.8673 & 0.2235
			& 28.5103 & 0.8503 & 0.2283 \\
			
			\rowcolor{gray!15}
			FastGS+Ours
			& \textbf{26.4326} & \textbf{0.8747} & \textbf{0.1775}
			& \textbf{29.6358} & \textbf{0.8705} & \textbf{0.2011} \\
			
			\hline
			FasterGS
			& 26.8850 & 0.8940 & 0.1825
			& 30.1022 & 0.8971 & 0.1634 \\
			
			\rowcolor{gray!15}
			FasterGS+Ours
			& \textbf{28.5700} & \textbf{0.9115} & \textbf{0.1555}
			& \textbf{31.7422} & \textbf{0.9217} & \textbf{0.1312} \\
			\hline
			\hline
		\end{tabular}%
	}
\end{table}

\subsection{Parameter and Nonlinear Mapping Analysis}

\begin{figure}[htbp]
	\centering
	\includegraphics[width=\columnwidth]{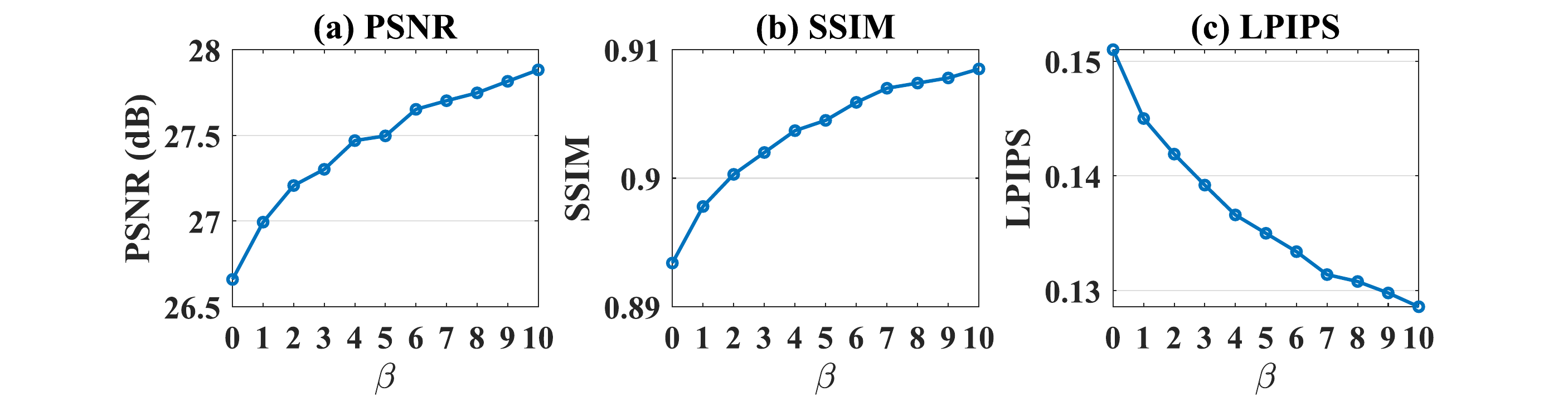}
	\caption{Effect of the weighting parameter $\beta$ on T\&T.}
	\Description{Three line plots showing how PSNR, SSIM, and LPIPS change when beta increases from 0 to 10.}
	\label{fig:beta_sweep}
\end{figure}

We first study the effect of the weighting parameter $\beta$ under the linear Laplacian-guided setting. As shown in Fig.~\ref{fig:beta_sweep},
larger $\beta$ strengthens the Laplacian guidance and improves performance within the evaluated range. Among the tested candidates, $\beta=10$
performs best and is used for the following nonlinear mapping analysis.

We then compare the five mappings in Fig.~\ref{fig:nonlinear_mapping} under the selected $\beta$. Table~\ref{tab:nonlinear_mapping} reports the
results on T\&T, with C1 as the linear reference. C3 achieves the best overall performance and improves PSNR by 0.2467 dB over C1, showing that
nonlinear response shaping benefits Laplacian-guided Gaussian optimization.

\begin{table}[htbp]
	\centering
	\caption{Nonlinear mapping analysis on T\&T. C1 is the linear reference.}
	\label{tab:nonlinear_mapping}
	\scriptsize
	\setlength{\tabcolsep}{4.5pt}
	\renewcommand{\arraystretch}{1.10}
	\resizebox{\columnwidth}{!}{%
		\begin{tabular}{c|cc|cc|cc}
			\hline
			\hline
			Curve 
			& PSNR$\uparrow$ & $\Delta$PSNR 
			& SSIM$\uparrow$ & $\Delta$SSIM 
			& LPIPS$\downarrow$ & $\Delta$LPIPS \\
			\hline
			C1 
			& 27.9494 & +0.0000 
			& 0.9079 & +0.0000 
			& 0.1291 & +0.0000 \\
			
			C2 
			& 27.7214 & -0.2280 
			& 0.9059 & -0.0020 
			& 0.1330 & +0.0039 \\
			
			C3 
			& \textbf{28.1961} & \textbf{+0.2467} 
			& \textbf{0.9092} & \textbf{+0.0013} 
			& \textbf{0.1269} & \textbf{-0.0022} \\
			
			C4 
			& 27.7581 & -0.1913 
			& 0.9069 & -0.0010 
			& 0.1309 & +0.0018 \\
			
			C5 
			& 28.0566 & +0.1072 
			& 0.9073 & -0.0006 
			& 0.1293 & +0.0002 \\
			\hline
			\hline
		\end{tabular}%
	}
\end{table}

\subsection{Visual Comparison}

\begin{figure}[htbp]
	\centering
	\setlength{\tabcolsep}{1.5pt}
	\renewcommand{\arraystretch}{1.0}
	
	\def\visimgw{0.24\columnwidth}
	\def\visimg#1{\includegraphics[width=\visimgw]{#1}}
	\def\vishead#1{\makebox[\visimgw][c]{\scriptsize\textbf{#1}}}
	
	\resizebox{\columnwidth}{!}{%
		\begin{tabular}{cccc}
			\vishead{3DGS~\cite{3DGS}} &
			\vishead{EGGS~\cite{EGGS}} &
			\vishead{LEGS} &
			\vishead{GT} \\
			
			\visimg{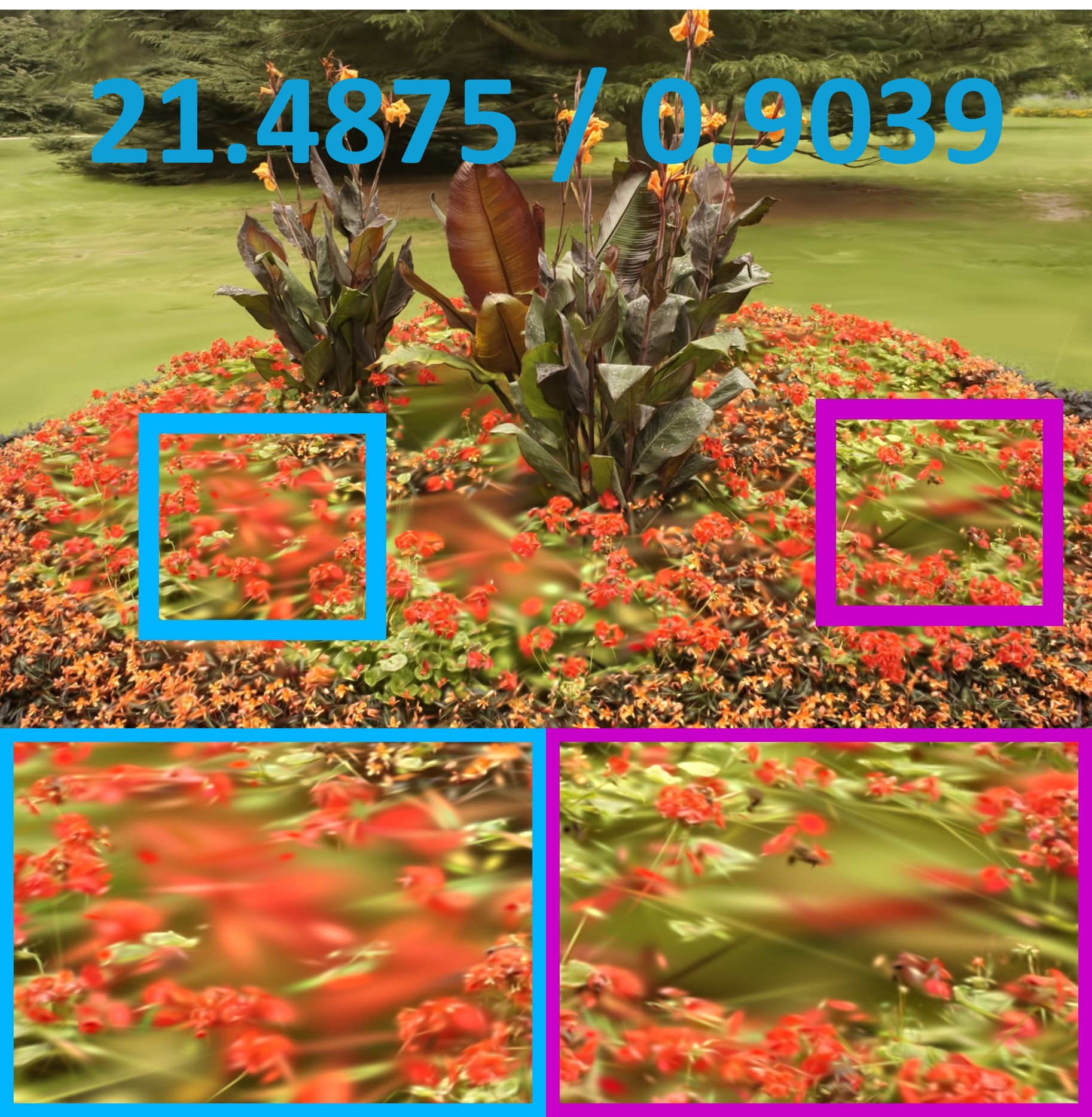} &
			\visimg{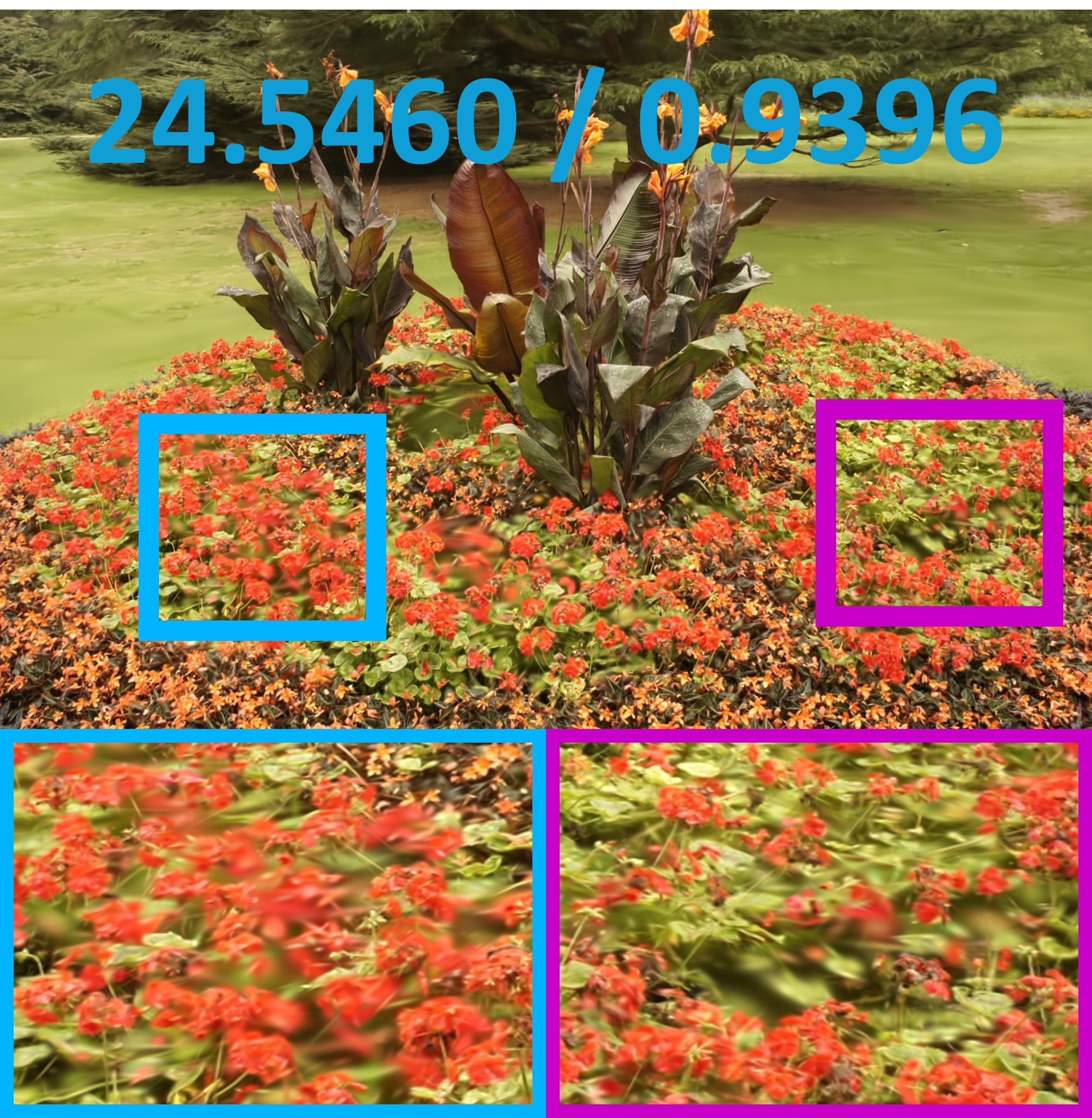} &
			\visimg{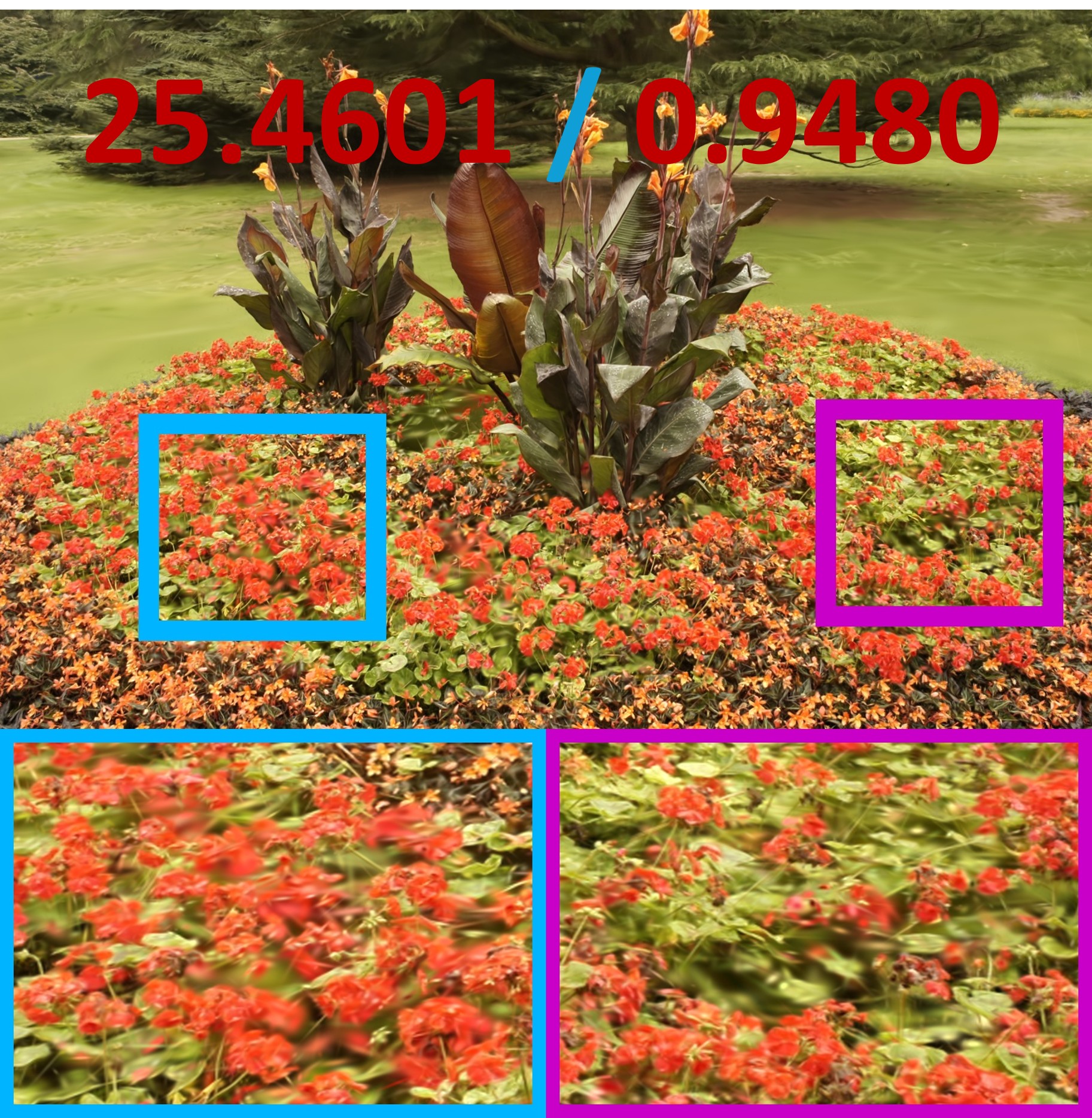} &
			\visimg{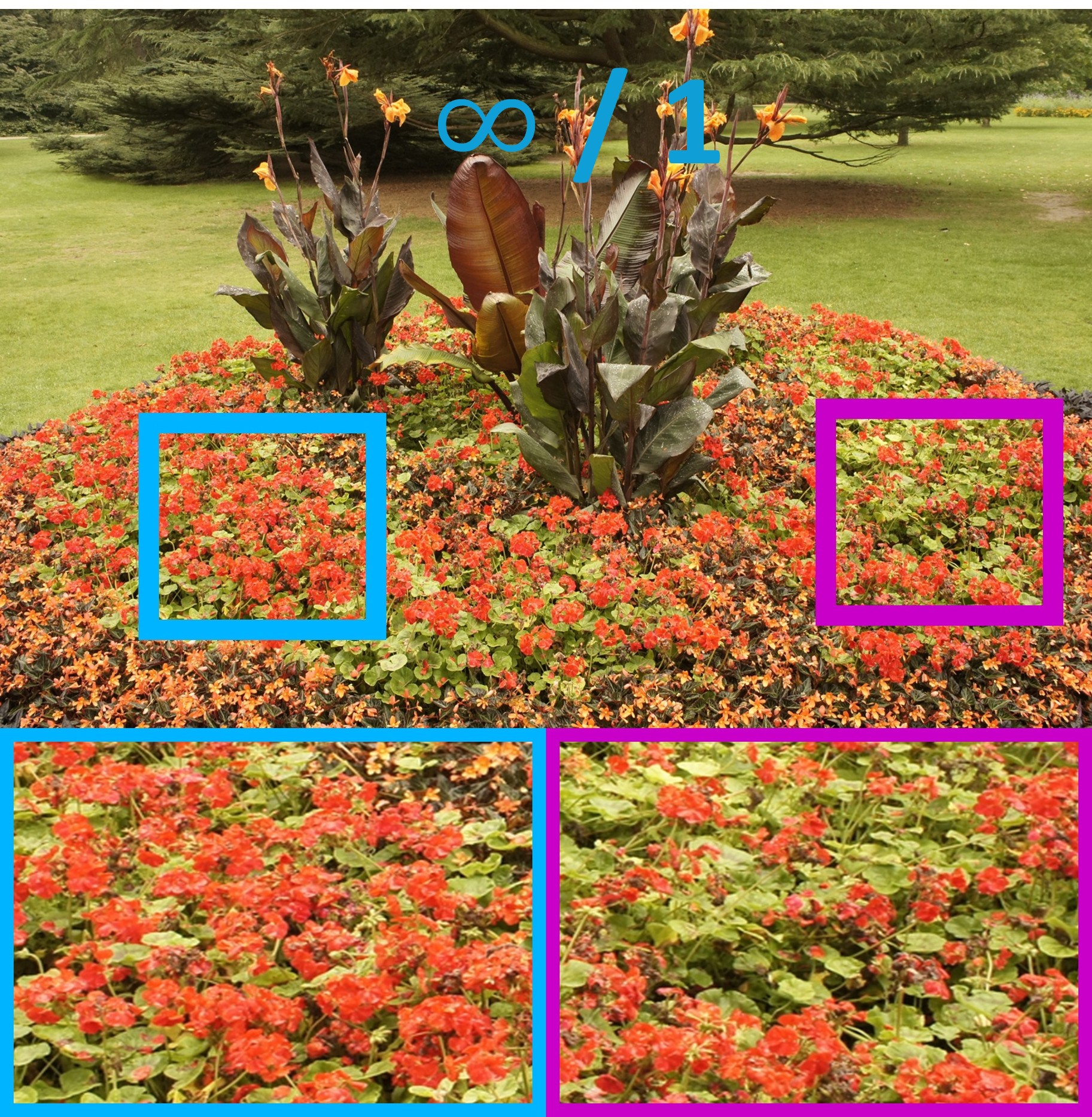} \\
			
			\visimg{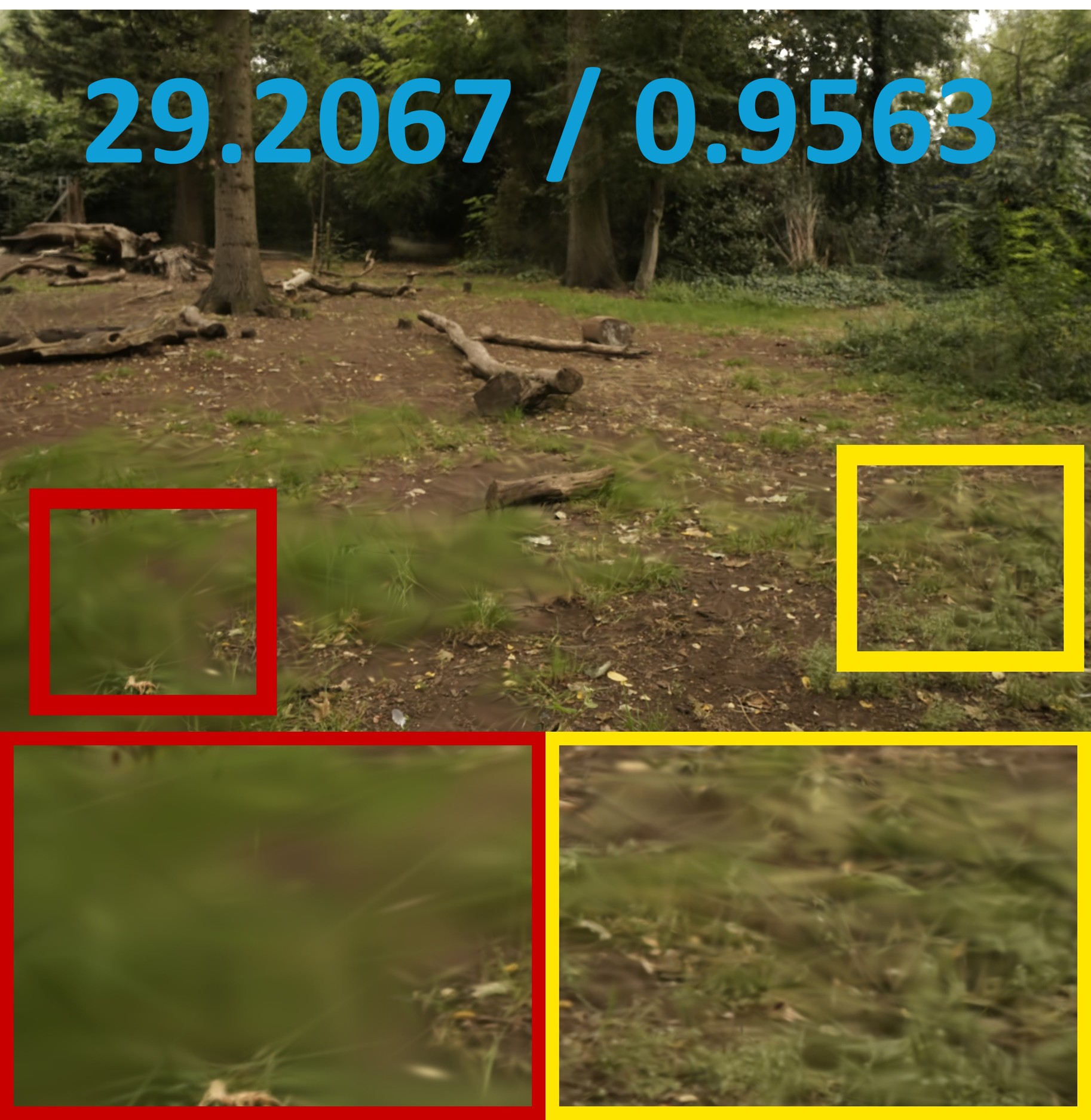} &
			\visimg{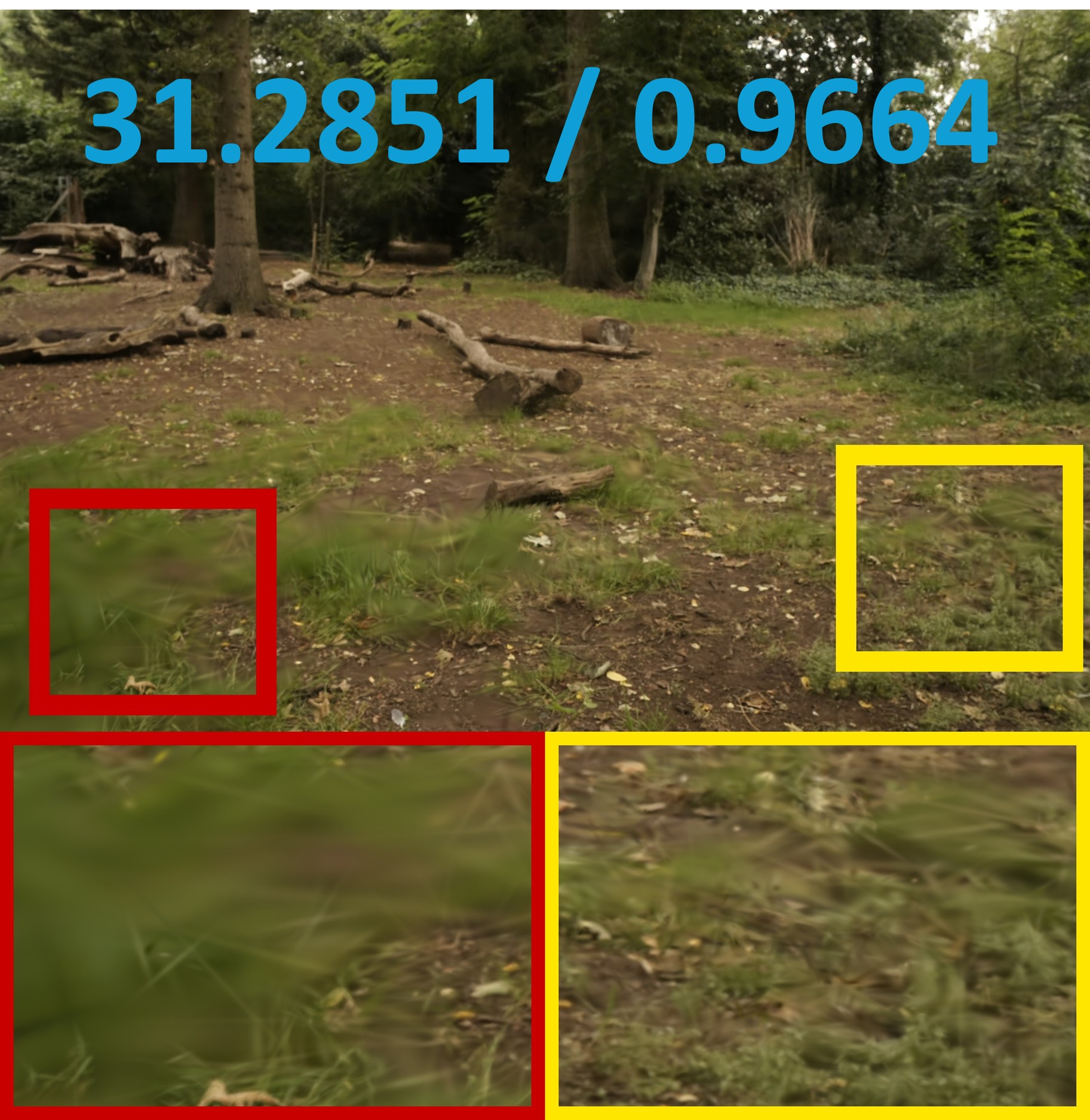} &
			\visimg{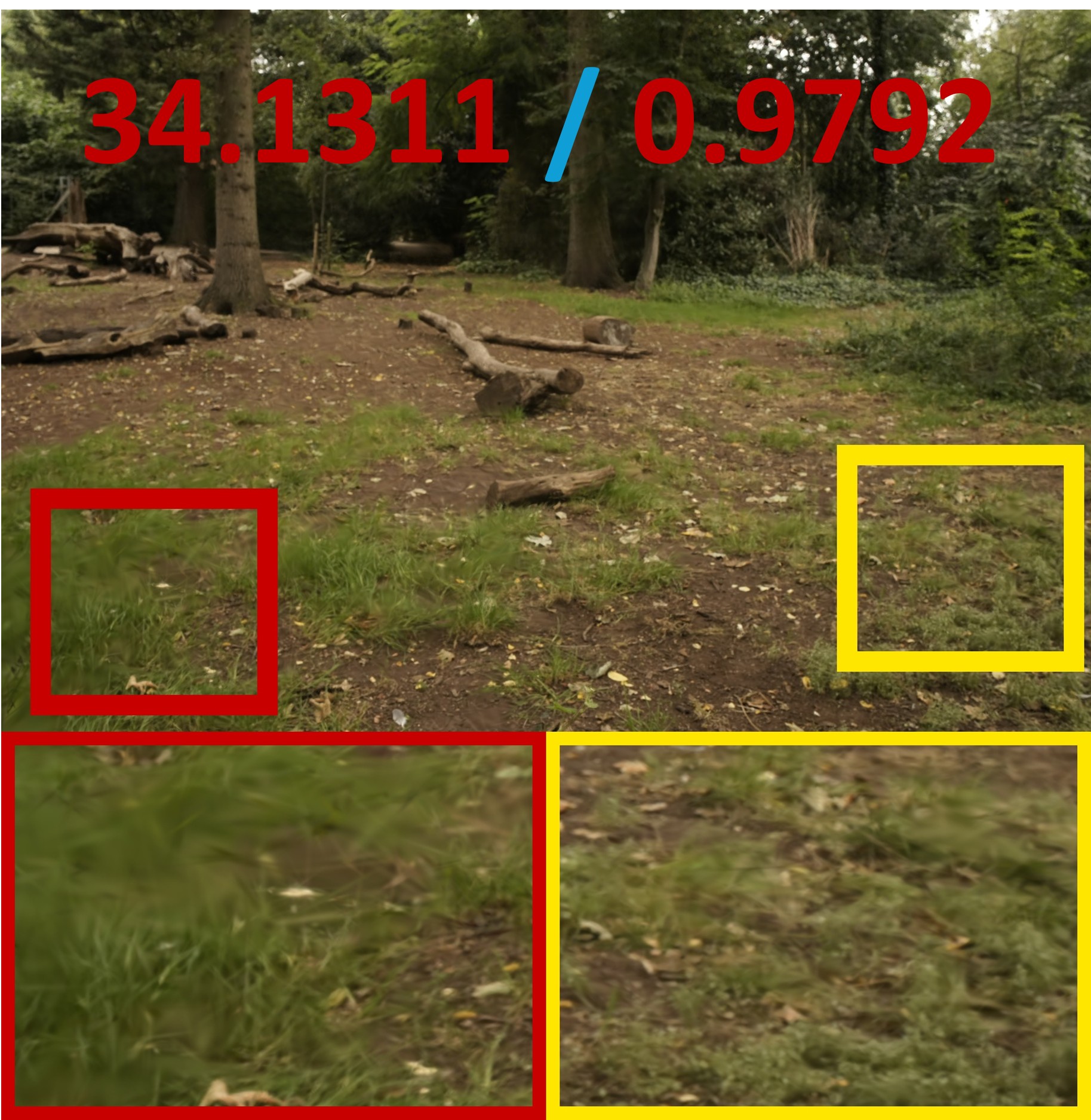} &
			\visimg{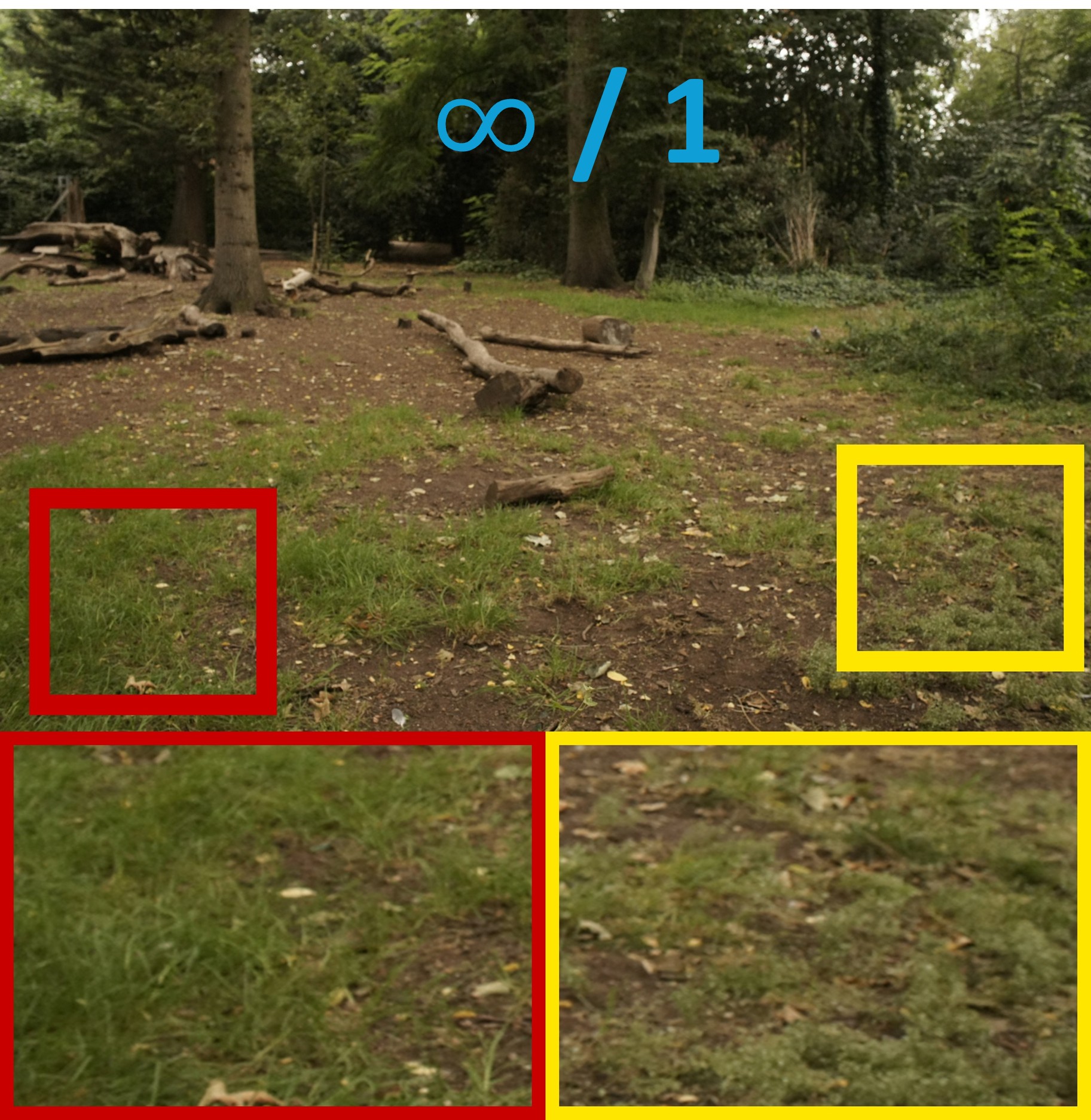}
		\end{tabular}%
	}
	\caption{Visual comparison on representative scenes.}
	\Description{A visual comparison showing rendered results from 3DGS, EGGS, LEGS, and the ground truth on two representative scenes.}
	\label{fig:visual_comparison}
\end{figure}

Fig.~\ref{fig:visual_comparison} shows visual comparisons on representative scenes. Compared with 3DGS and EGGS, LEGS better preserves contours, corners, and fine local structures. This agrees with the quantitative results and indicates that second-order Laplacian structural guidance helps the optimization allocate more attention to visually important regions.

\section{Conclusion}

In this paper, we proposed LEGS, a Laplacian-Enhanced Gaussian Splatting method with a nonlinearly weighted loss. LEGS introduces second-order Laplacian structural guidance and nonlinear response-to-weight mapping to improve structure-aware Gaussian optimization, while keeping the original Gaussian representation, renderer, and optimization pipeline unchanged.

Experiments on T\&T and Mip-NeRF360 show that LEGS improves reconstruction quality over 3DGS, EGGS, and the linear Laplacian-weighted variant. Results on FastGS and FasterGS further show its generality as a loss-level extension for Gaussian Splatting pipelines. In future work, we will explore more adaptive structural weighting strategies and extend Laplacian-enhanced supervision to more efficient and compact Gaussian representations~\cite{StruGS,Opt3DGS,IDGS}. More broadly, LEGS can be applied to a wide range of radiance-field and real-time 3D applications where local structural information is important~\cite{MC,MSBM,osbf,FHQ,3DGEER,FPGAOSBF,CCMC,FESW,CCC,zhou2025gradient,MSE,WMC,CSDF,LWGC,CFE,STR}.

\section{Acknowledgments}

This work was supported in part by National Natural Science Foundation of China under Grant 12471502, CAS Hundred Talents Program, and Science and Technology Development Plan Project of Jilin Province, China under Grant 20260204053YY. Contact:gong.ai@qq.com.

\bibliographystyle{ACM-Reference-Format}
\bibliography{sample-base}

\end{document}